\documentclass[9pt]{extarticle}
\usepackage{spconf,amsmath,graphicx}
\usepackage{placeins}
\usepackage{epstopdf}
\usepackage{latexsym}
\usepackage{amssymb}
\usepackage{epsfig, psfrag}
\usepackage{amsthm} 
\usepackage{graphics,color}
\usepackage{amsmath}
\usepackage{enumerate}
\usepackage{algorithm2e}
\usepackage{url}
\usepackage{xspace}
\usepackage{xcolor}
\usepackage{soul}
\usepackage[colorinlistoftodos]{todonotes}

\newcommand{\ignore}[1]{}
\newcommand{\add}[1]{{\color{blue}{#1}}}

\def\argmax{\mathop{\rm argmax}}

\title{Automatic Grammar Augmentation for robust voice command recognition}
\name{Author(s) Name(s)}
\address{Author Affiliation(s)}

\name{Yang Yang$^{ \dagger}$, Anusha Lalitha $^{\star}$, Jinwon Lee$^{\dagger}$, Chris Lott$^{\dagger}$ }
\address{$^{\dagger}$ Qualcomm Research, San Diego \\$^{\star}$ Department of Electrical and Computer Engineering, University of California San Diego}

\begin{document}
%
\maketitle
\begin{abstract}

This paper proposes a novel pipeline for automatic grammar augmentation that provides a significant improvement in the voice command recognition accuracy for systems with small footprint acoustic model (AM). The improvement is achieved by augmenting the user-defined voice command set, also called grammar set, with alternate grammar expressions. For a given grammar set, a set of potential grammar expressions (candidate set) for augmentation is constructed from an AM-specific statistical pronunciation dictionary that captures the consistent patterns and errors in the decoding of AM induced by variations in pronunciation, pitch, tempo, accent, ambiguous spellings, and noise conditions. Using this candidate set, greedy optimization based and cross-entropy-method (CEM) based algorithms are considered to search for an augmented grammar set with improved recognition accuracy utilizing a command-specific dataset. Our experiments show that the proposed pipeline along with algorithms considered in this paper significantly reduce the mis-detection and mis-classification rate without increasing the false-alarm rate. Experiments also demonstrate the consistent superior performance of CEM method over greedy-based algorithms. 
\end{abstract}

\vspace{-0.5cm}
\section{Introduction}
\label{sec:intro}
\vspace{-0.2cm}

Voice UI is becoming ubiquitous for all types of devices, from smartphones to automobiles. Although we have seen substantial improvement in speech recognition accuracy reported in the literature since the advent of deep neural network based solutions \cite{DS1, DS2, Wav2Letter2016,RnnTransducer2012},  designing robust voice UI system for low memory/power footprint embedded devices without a cloud-based back-end still remains a challenging problem. 
Compared to its cloud-based counterpart, on-device inference, despite being limited by computation power, memory size, and power consumption, remains appealing for several reasons: (i) there are less privacy concerns as user voice data need not  be uploaded to the cloud; (ii) it reduces the latency as it does not involve network access delay; (iii) its usage is not restricted by internet availability, and can be applied in devices with no built-in communication module.

In this work, we focus on improving the recognition accuracy of on-device voice UI systems designed to respond to a limited set of pre-defined voice commands. Such voice UI systems are commonly used in modern IoT/embedded devices such as bluetooth speaker, portable camcorder, hearables, home appliances, etc. Specially, we assume a fixed audio front-end and only look at the pipeline of mapping acoustic features to voice commands.

\begin{figure}[t]
\begin{center}
\includegraphics [width=0.45\textwidth]{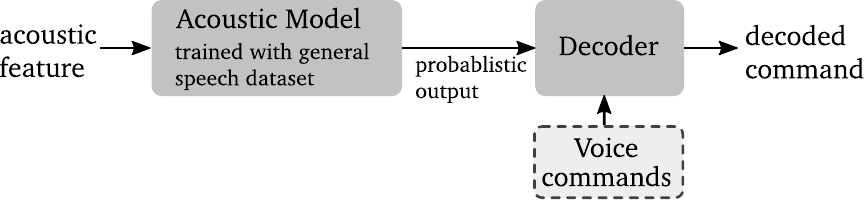}
\caption{
Voice command recognition pipeline
}\vspace{-0.7cm}
\label{fig:two_approaches}
\end{center}
\end{figure}

As illustrated in Fig.~\ref{fig:two_approaches}, we focus on the voice command recognition system composed of an acoustic model (AM) encoder that converts the acoustic features into phoneme/grapheme-based probabilistic output, followed by a decoder (e.g., FST) that maps the probabilistic output from AM to one of the voice commands. State of the art acoustic model utilizes either CTC \cite{AG_CTC_ICML2006}, RNN-transducer \cite{RnnTransducer2012}, or Attention Model \cite{AttentionModelBengio} (see \cite{GoogleNvidiaSummary,BaiduSummary} for a good summary). They generate probabilistic outputs, which are fed to a decoder that generates the posterior probability of the corresponding phoneme or grapheme label. Even though these model architectures and training methodologies lead to satisfactory and even super-human transcription accuracy, the best models obtained are often too large for their deployment in small portable devices, e.g., even the smallest model considered in \cite{DS2-arxiv} (Table 11 therein) has 18M parameters. 

In this work, we utilize a 211K parameter unidirectional-RNN-based acoustic model trained with CTC criterion using Librispeech and a few other datasets, which output probabilities on grapheme targets. Due to the small model size, its transcription accuracy is low: the greedy decoding word-error-rate (WER) without any language model is $48.6\%$ on Libri-speech test-clean dataset. Hence, one of the challenges addressed by our work is, given a small acoustic model trained with general speech dataset, how can one improve the command recognition accuracy utilizing limited command-specific data. Such small footprint AMs have been considered for keyword detection in~\cite{DBLP:conf/interspeech/SainathP15} and~\cite{6854370}. Our work extends these by improving the command command recognition accuracy with a small footprint AM.

\begin{table}[b]
\begin{center}\vspace{-0.4cm}
  \begin{tabular}{  l | l }
    \hline
    AM greedy decoding & Ground truth  \\ \hline\hline
    \small the {\bf recter pawsd} and {\bf den} & \small the rector paused and then\\ 
    \small shaking his {\bf classto} hands & \small shaking his clasped hands \\ 
    \small  before him went on & \small before him went on \\ \hline
    \small {\bf tax} for {\bf wone o thease}  &\small  facts form one of these \\
    \small  and {\bf itees he} other & \small  and ideas the other   \\
    \hline
  \end{tabular}\vspace{-0.4cm}
\end{center}
\caption{Greedy decoding samples from the acoustic encoder. 
Word errors are labeled in bold.}
\label{table:decoding_samples}\vspace{-0.3cm}
\end{table}

In Table~\ref{table:decoding_samples}, we list a few samples of the greedy decoding results from the 211K parameter acoustic model. It is worth noting that even though the word-error-rate is high, the error that it makes tends to be a phonetically plausible rendering of the correct word \cite{DS1}. Running through a large dataset, we also observe that the error patterns tend to be consistent across different utterances. This leads to a useful insight: for the recognition of a limited set of voice commands (a.k.a. grammar of the decoder), one could improve recognition accuracy by adding variations that capture common and consistent errors from the acoustic model to original command set. We define grammar as a set of valid voice commands (e.g., the grammar can be $\{$\emph{play music}, \emph{stop music}$,\ldots\}$) and we refer to this technique of adding variations to the original grammar as \emph{grammar augmentation}. Effective grammar augmentation is the focus of this work.

The main contribution of this paper is the design of effective grammar augmentation framework which provides significant improvement over the baseline system. Next, we highlight our main contributions in detail: (a) For any given set of original voice commands, we propose the design of a candidate set of all grammar variations which captures the consistent errors for a given AM (b) We propose a technique for fast evaluation of command recognition accuracy along with false-alarm and mis-detection rate for any augmented grammar set and finally (c) We devise various algorithms to automatically identify an improved augmented grammar set by suitably adding variations from the candidate set to the original grammar.

Our novel pipeline using the above techniques is illustrated in Fig.~2. The rest of the paper is organized as the following: In Section~\ref{sec:pipeline}, we give an overview of the proposed grammar augmentation pipeline and dive into the generation of a candidate set and fast grammar evaluation techniques. In Section~\ref{sec:aug_search} algorithms via greedy optimization and CEM algorithm are utilized to automate the grammar augmentation process. The experiment results are presented in Section~\ref{sec:experiment} and we  discussion on future directions in Section~\ref{sec:conclusion}.

\begin{figure*}[t]
\begin{center}
\includegraphics [width=1.00\textwidth]{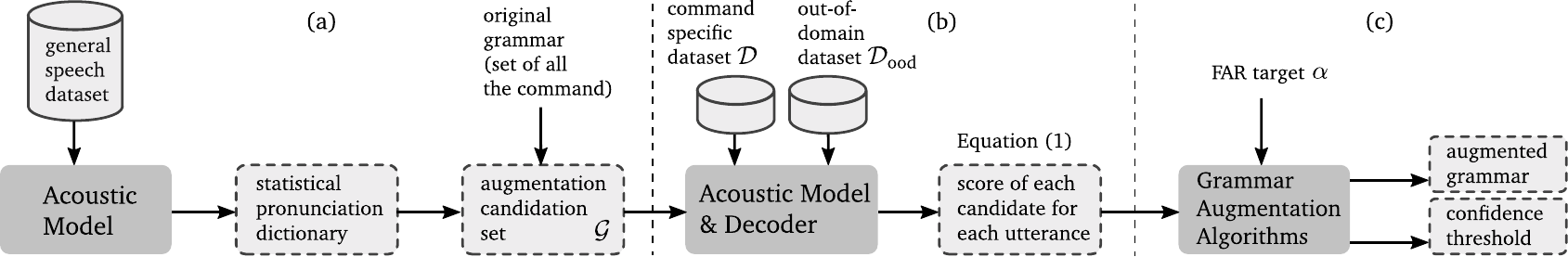}\vspace{-0.2cm}
\caption{Grammar augmentation pipeline. }\vspace{-0.4cm}
\end{center}
\label{fig:augmentation_pipeline}
\end{figure*}

\vspace{-0.2cm}
\section{pipeline for automatic grammar augmentation}
\vspace{-0.2cm}
\label{sec:pipeline}

Our AM is trained with CTC loss \cite{AG_CTC_ICML2006}, and can thus assign a posterior probability $\mathbb{P}_\text{CTC}(g|u)$ for each command $g$ in a command set, for an input utterance $u$. For a given test utterance, our system picks the command with the highest probability, or rejects the utterance if the highest probability is below a pre-defined confidence threshold (see Section~\ref{subsec:eval}) \cite{EESEN}\cite{Kanda2016MaximumAP}.

Command decoding errors happen if the AM output deviates from the ground truth to the extent that it can no longer successfully discriminate against other grammar sequences. The idea behind grammar augmentation is to restore the discriminative power of the acoustic model by including in the grammar the sequence variations that capture pronunciation variations or consistent AM error patterns. To do that, we begin with generation of a candidate set containing meaningful variations.

\vspace{-0.3cm}
\subsection{AM-specific statistical pronunciation dictionary}
\vspace{-0.1cm}

\label{subsec:pronun_dict}

The augmentation candidates should ideally capture consistent error patterns from the AM, induced by variations in pronunciation, pitch, tempo, accent, ambiguous spellings, or even inherent mistakes made by the AM. 
For example, if any command includes words that have homophones, then it is necessary to consider adding those homophones into the grammar. 
To capture these word-level variations, we introduce a novel concept named \emph{AM-specific statistical pronunciation dictionary}, obtained by the following steps: First, we run the AM through a large general speech dataset (e.g., the training set of AM). For each utterance, we obtain its greedy decoding sequence by outputting the character with the maximum probability at each time frame, followed by the CTC squashing function \cite{AG_CTC_ICML2006} to collapse repeated output graphemes and remove blanks. 
Given that most utterances from a general speech dataset correspond to a sentence rather than a single word, we use Levenshtein algorithm to find the minimum-edit-path of the ground-truth to the decoding, and by doing so obtain a mapping of each word to its corresponding maximum probability decoding. For each word, we gather the statistics regarding the frequencies of its maximum-probability decoding outputs. Here we sample a few entries from the dictionary obtained using our 211K-parameter AM:

\begin{center}
\vspace{-0.1cm}
  \begin{tabular}{  l  l  |  l  l  |  l l }
    \hline
     \multicolumn{2}{c}{ set}		&\multicolumn{2}{c}{ pause}		&\multicolumn{2}{c}{ two}	
     \\ \hline\hline
    \small set & \small 32.2\%		&\small pause & \small 15.7\%	&\small to & \small 53.3\%	
    \\ \hline
    \small said & \small 16.6\% 	&\small pose & \small 14.9\%	&\small two & \small 34.7\%	
    \\ \hline
    \small sat & \small 11.4\% 	&\small pase & \small 7.68\%	&\small do & \small 1.0\%	
    \\ \hline
    \small sait & \small 8.15\%	&\small porse & \small 7.31\%	&\small tu & \small 0.7\%	
    \\ \hline
    \small sed & \small 4.71\% 	&\small pas & \small 7.31\%		&\small too & \small 0.3\%
    \\ \hline
  \end{tabular}
  \vspace{-0.1cm}
\end{center}

\subsection{Candidate set for grammar augmentation}
\label{subsec:candidate_set}

Utilizing this statistical dictionary, we build a candidate set containing potential grammar variations by repeatedly replacing each word in the original grammar by its top-$k$ likely max-decoding outputs. Consider a voice UI application for a small bluetooth player, one could have the following five commands forming the original grammar.
\begin{center}
  \begin{tabular}{@{} l | l  | l @{}}
    \hline
    \small command&\small original& \small candidate set for\\ 
    \footnotesize  ($\mathcal{C}$)& \small grammar\footnotemark & \small  grammar augmentation ($\mathcal{G}$) \\\hline\hline
    \footnotesize play music&\footnotesize play music& \footnotesize pla music, ply music, play mesic, \ldots 
    \\ \hline
    \footnotesize stop music&\footnotesize stop music& \footnotesize stap music, stup music, stup mesic, \ldots 
    \\ \hline
    \footnotesize pause music&\footnotesize pause music& \footnotesize pose music, pase mesic, pause mesic, \ldots 
    \\ \hline
    \footnotesize previous song&\footnotesize previous song& \footnotesize previs song, previous son, \ldots
    \\ \hline
    \footnotesize next song&\footnotesize next song& \footnotesize nex song, lext song, nex son, \ldots 
    \\\hline
  \end{tabular}
\end{center}
\footnotetext{Here we assume that AM is trained with grapheme as target, and as a result the grammar is exact the same as the command. Note that the same grammar augmentation pipeline introduced here can be applied to AM trained with phoneme target as well, in which case the grammar is a set of phoneme sequences, and the statistical pronunciation dictionary contains variations of each word in phoneme representation.}
By looking up in the statistical dictionary the words contained in the original grammar, one can form an array of alternate expressions for the original commands as shown above. For each command, the set of candidates is the cartesian product of the top-$k$ decoding list from the statistical pronunciation dictionary for each word in the command. The value of $k$ can be different for different words, and is chosen to capture at least a certain fraction of all the variations.

\vspace{-0.2cm}
\subsection{Evaluation of command recognition accuracy}
\vspace{-0.1cm}
\label{subsec:eval}

Let us denote the set of commands as $\mathcal{C}$, the set of all grammar candidates $\mathcal{G}$, and the mapping function from $\mathcal{G}$ to $\mathcal{C}$ as $f$. A grammar $G$ is a subset of $\mathcal{G}$. For the purpose of evaluating the recognition accuracy of any grammar, we need a command-specific dataset containing audio waveforms and the corresponding target commands. We denote such dataset as $(u,t)\in\mathcal{D}$ with $u$ and $t$ denoting an utterance and its corresponding target command.  To evaluate the false alarm rate, we also need an out-of-domain dataset $u\in\mathcal{D}_{\text{ood}}$ that contains a set of utterances that do not correspond to any of the commands. 

As mentioned before, the acoustic decoder compares the posterior probabilities $\mathbb{P}_\text{CTC}(g|u)$ of all the grammar candidates $g$ included in grammar set $G \subset \mathcal{G}$ given the audio waveform $u$, and output the command $f(g^{\ast})$ where $g^{\ast}=\argmax_{g\in G}\mathbb{P}_\text{CTC}(g|u)$. 
This calculation is done by running a forward-only dynamic programming algorithm on the AM output. In order to avoid having to repeat the calculation of the probability scores for every choice of grammar set $G\subseteq \mathcal{G}$, we pre-compute and store the probability scores for all the candidate grammar, and all the utterances in both command-specific dataset $\mathcal{D}$ and out-of-domain dataset $\mathcal{D}_\text{ood}$. Precisely, as a pre-processing steps of the grammar augmentation search algorithm, we obtain the following probability scores:
\begin{align}
\mathbb{P}_\text{CTC}(g|u),\hspace{0.5cm}\forall g\in\mathcal{G}, \forall u\in\mathcal{D}\cup\mathcal{D}_{\text{ood}}\label{eq:pre-compute}.
\end{align}
To achieve a false alarm rate (FAR) target of $\alpha$, the confidence threshold for the probability score can be computed as below, 
\begin{align}
\tau(G,\alpha) = \min_{\tau} {\Bigg\{}  \tau: \frac{   {\big|}{\big\{}u\in \mathcal{D}_{\text{ood}}: \underset{g\in G}{\max} \mathbb{P}_\text{CTC}(g|u)>\tau {\big\}}{\big|} }{ \left|\mathcal{D}_{\text{ood}}\right|   } < \alpha  {\Bigg\}}\notag.
\end{align}
The decoded command for an utterance $u$ is
\begin{align}
d(G, \alpha, u) = \left\{
\begin{array}{l }
\phi, \hspace{1.3cm}\text{ if }\max_{g\in G}\mathbb{P}_\text{CTC}(g|u) < \tau(G,\alpha),\\
\underset{c\in\mathcal{C}}{\argmax} \underset{g\in G, f(g)=c}{\max} \mathbb{P}_\text{CTC}(g|u),\hspace{0.8cm}\text{otherwise.}
\end{array}
\right.\notag,
\end{align}
$\phi$ denotes decoding being out-of-domain. 

With a fixed false-alarm rate, there are two types of error event: mis-detection and mis-classification. Mis-detection refers to the case where a voice command is issued but not detected (i.e., decoded as being out-of-domain), whereas mis-classification happens where a voice command is issued and detected, but the wrong command is decoded. Precisely, the mis-detection-rate (MDR) and the mis-classification-rate (MCR) are defined as below
\begin{align}
\text{MDR}(G,\alpha) =& 
|  \left\{ (u,t)\in \mathcal{D}: d(G,\alpha, u)=\phi \right\}  | / |\mathcal{D}|
,\notag\\
\text{MCR}(G,\alpha) =& 
|  \left\{ (u,t)\in \mathcal{D}: d(G,\alpha, u)\not\in\{\phi, t\} \right\}  |/|\mathcal{D}|
.\notag
\end{align}

\section{Augmentation search algorithms}
\label{sec:aug_search}
\vspace{-0.2cm}

The grammar augmentation algorithms we consider search for the grammar set $G$ among all subsets of a candidate set $\mathcal{G}$ that minimizes a weighted sum of the mis-detection-rate and mis-classification-rate with a fixed false-alarm target $\alpha$, 
\begin{align}
\min_{G\subseteq \mathcal{G}} \text{MCR}(G, \alpha)  + \beta\text{MDR}(G, \alpha).\label{eq:objective}
\end{align}
Here the weight factor $\beta$ controls the significance of mis-detection versus mis-classification. 
Since we pre-compute the probabilities as shown in Equation~\eqref{eq:pre-compute}, for each grammar $G\subseteq\mathcal{G}$ the objective function can be evaluated without invoking the AM, which significantly speeds up the search algorithms.

It is important to note that adding candidate to the grammar does not always improve performance: (i) 
With a fixed false-alarm target, adding more candidates only increase the confidence threshold $\tau(G,\alpha)$, which could potentially result in degraded mis-detection rate. 
(ii) distinguishability of the commands has a complex inter-dependency, hence adding grammar candidate for one command may reduce the recognition rate of other commands, as it may alter the classification boundary amongst the set of commands.

\vspace{-0.4cm}
\subsection{Augmentation via greedy optimization methods}
\vspace{-0.2cm}

We consider the following three methods based on greedy optimization:

\underline{\emph{Naive greedy search}:} Start with the original grammar, iteratively go through all the candidates from $\mathcal{G}$. In each iteration, add the candidate that best improves the objective function and update the confidence threshold to maintain target FAR, until no candidate can improve further.

\underline{\emph{Greedy search with refinement}:} This algorithm is similar to greedy search except for every time a candidate is added to the grammar, we remove those candidates among the remaining ones which contain the added candidate as a subsequence. For example, for \emph{pause music} command, if candidate \emph{pose music} is added to the grammar, then \emph{porse music} is removed from subsequent iterations. Trimming the candidate set in this manner increases the diversity of variations in the grammar.

\underline{\emph{Beam-search:}}  In each iteration a list of $l$ best grammar sets is maintained. This degenerates to the naive greedy algorithm when $l=1$.

\vspace{-0.4cm}
\subsection{Augmentation via cross entropy method (CEM)}
\vspace{-0.2cm}

Cross entropy method (CEM) is a widely used combinatorial optimization algorithm and has been successfully applied in some reinforcement learning problems~\cite{tetris_paper, de_boer_tutorial}. The main idea is rooted from rare event sampling, for which the algorithm tries to minimize the KL divergence between a proposed sampling distribution and the optimal zero-variance importance sampling distribution \cite{de_boer_tutorial}.  Going back to the grammar augmentation objective function in Equation~\eqref{eq:objective}, the search space is the power set of the candidate set $\mathcal{G}$, which can be represented by $\{0,1\}^{|\mathcal{G}|}$, with each grammar choice represented by a $|\mathcal{G}|$-dimensional binary vector. 

Applying the idea of CEM, we start with an initial probability distribution on $\{0,1\}^{|\mathcal{G}|}$, and iteratively tune its parameter so that it assigns most of the probability mass in the region towards the minimization of the objection function. In our design, the distribution on this discrete space is induced by the sign of a $|\mathcal{G}|$-dimensional independent Gaussian distributions, parameterized by their mean and variance in each dimension.
For each iteration, we start with a population of $s$ samples from the current distribution, each representing a feasible candidate choice. We evaluate the objective function of $\text{MDR}+\beta\text{MCR}$ for each of sample candidate choice, and keep the best $\gamma$ fraction. We then update the parameter of the distribution using the sample mean and variance of the top $\gamma s$ candidates (also called elite set), and iterate the procedure by obtaining $s$ samples from the updated distribution.

\vspace{-0.3cm}
\section{Experiments}
\vspace{-0.2cm}
\label{sec:experiment}

In this section, we present some experiments which illustrate the improvement that can obtained in recognition accuracy by applying our grammar augmentation algorithm. 
All the results are obtained with a dataset containing 5 commands: \emph{play music, pause music, stop music, next song} and \emph{previous song}. This dataset contains utterances with varying gender, pitch, volume, noise types and accents, and are split into training, validation, and testing datasets.
The training dataset is used to train the augmentation search algorithms to minimize the objective defined in~\eqref{eq:objective}. 
The validation dataset is used to compare performances of grammar sets obtained and decide which one to take. 
Finally, we report the results of the final grammar set on a test dataset. For the training objective function in Equation~\eqref{eq:objective}, we pick $\beta = 1$, in which case minimizing the sum of MDR and MCR is equivalent to maximizing the command success rate $1- \text{MCR}(G, \alpha)- \text{MDR}(G, \alpha)$.
A candidate set is obtained from running the 211K parameter AM with a 2000-hour dataset using steps discussed in Section~\ref{subsec:pronun_dict} and~\ref{subsec:candidate_set}.
 We consider 150 grammar candidates ($|\mathcal{G}|=150$) using our statistical pronunciation dictionary.


\subsection{Performance Evaluation}
\begin{figure}[!htb]
\begin{center}
\includegraphics [width=0.45\textwidth]{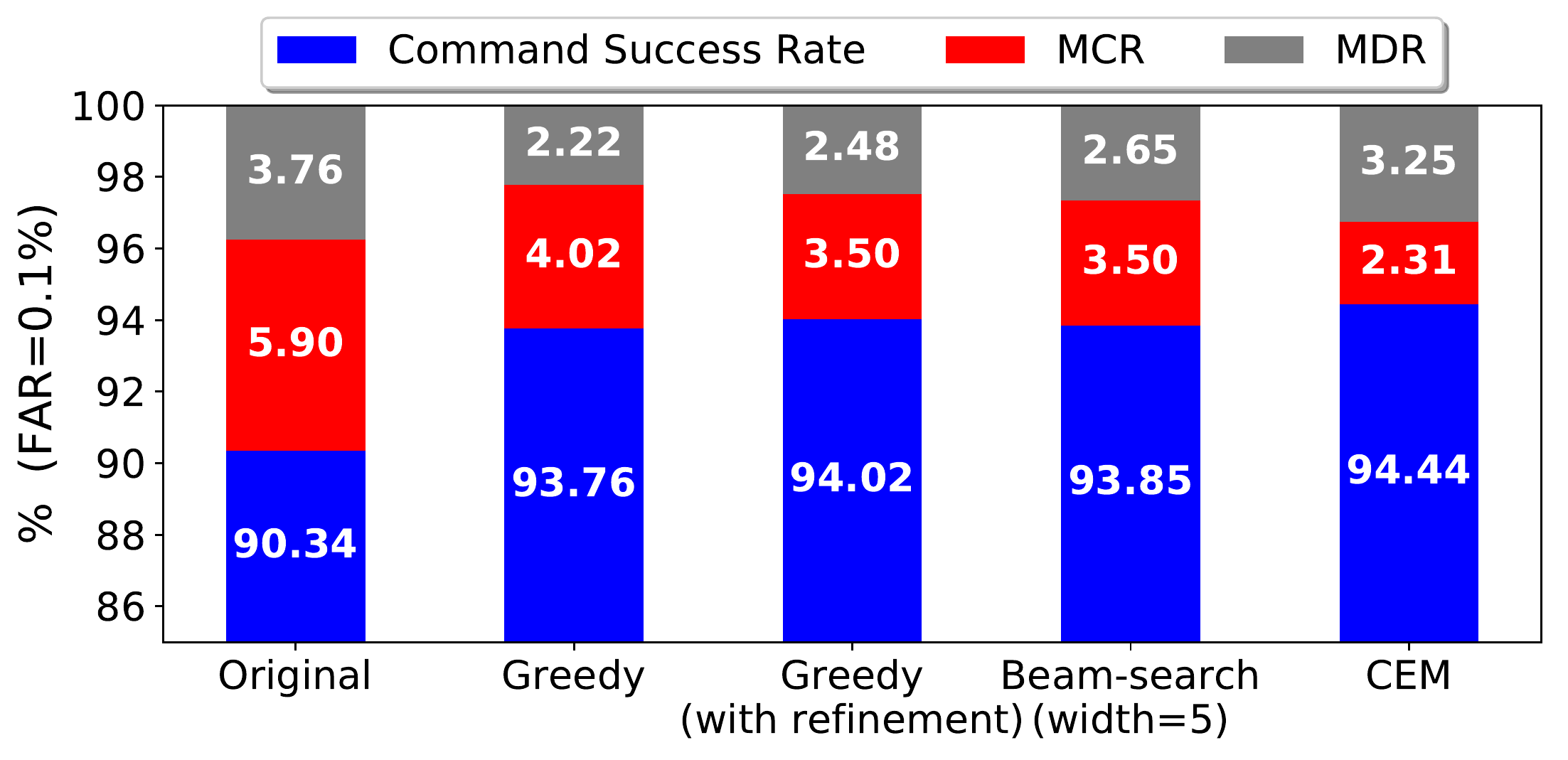}\vspace{-0.3cm}
\caption{Performance of grammar augmentation algorithms. 
}
\label{fig:isr_values}
\end{center}
\end{figure}

We analyze the grammar augmentation algorithms described in Section~\ref{sec:aug_search} with a fixed FAR target of $\alpha=0.1\%$ and compare the augmentation grammar output by each algorithm in terms. Fig.~\ref{fig:isr_values} shows the command success rate and the decomposition of the error in terms of mis-detection and mis-classification. Note that CEM algorithm provides most improvement in command success rate unlike greedy-optimization based algorithms which may commit to sub-optimal grammar sets early on. As discussed previously, adding more variations to the grammar set makes it more susceptible to mis-detection errors.
In fact, adding all 150 grammar expression reduces the command success rate to $80\%$ and increases the MDR to $13.76\%$. However, Fig~\ref{fig:isr_values} shows that performing augmentation in a principled manner can greatly reduce the mis-classification error without increasing the mis-detection errors. 

\vspace{-0.4cm}
\subsection{Complexity of Grammar Augmentation Algorithms}

\begin{figure}[!htb]
\begin{center}
\includegraphics [width=0.40\textwidth]{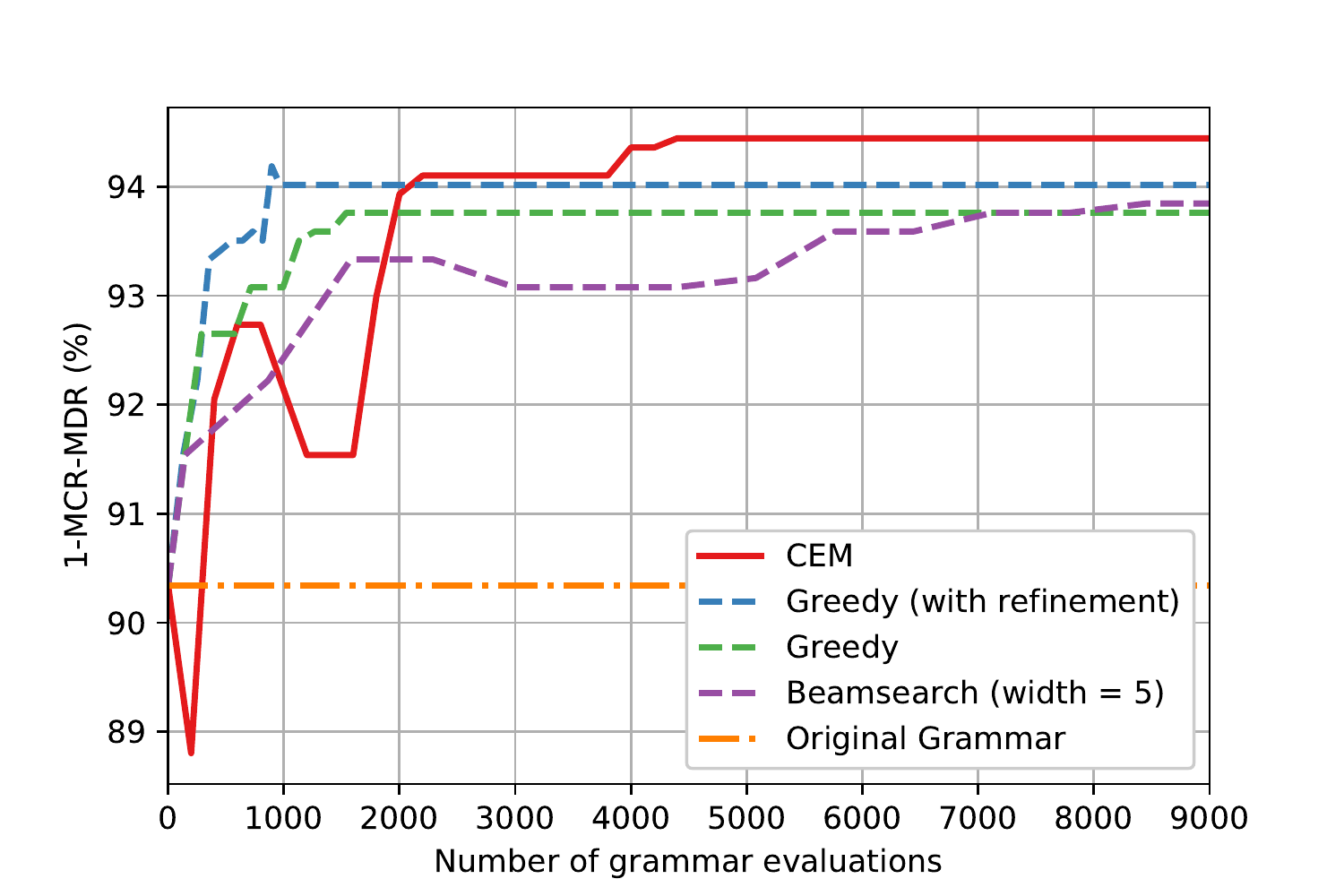}\vspace{-0.3cm}
\caption{Test dataset performance vs.~number of grammar evaluations}\vspace{-0.7cm}
\label{fig:complexity}
\end{center}
\end{figure}

We evaluate the complexity of the augmentation algorithms considered in Section~\ref{sec:aug_search}. The most computationally expensive step in implementing our augmentation algorithms is the evaluation of MCR and MDR for any candidate grammar set. Hence, we measure the complexity of our augmentation algorithms in terms of number of grammar evaluations needed to output their best augmented grammar set. Fig.~\ref{fig:complexity} illustrates the variation/improvement in command success rate (1-MDR-MCR) as the number of grammar evaluations increases. Note that CEM takes only marginally more evaluations while providing the maximum reduction in the sum of MCR and MDR. While beamsearch explores more and requires more grammar evaluation\add{,} it provides only marginally better improvement over naive greedy. The greedy algorithm refinement reaches its best performance in the least number of grammar evaluations. This suggests that incentivizing diversity over exploration may provide better improvement in command success rate and in fewer evaluations.



\subsection{Effect of Candidate Set Size on Performance}

So far we considered a candidate set size of 150 ($|\mathcal{G}|=150$). Next, we investigate the effect of varying the candidate set size on the performance of the augmentation algorithms. We vary the candidate size by varying the number of words $k$ we choose from the top-$k$ likely max-decoding outputs for every word in the statistical pronunciation dictionary. Hence, a larger candidate size captures a larger probability of max-decoding outputs. We repeat our experiments by altering the candidate set size from 25 to 150. Table~\ref{table:candidate_size} shows the performance the augmentation algorithms for various candidate set sizes. In particular, it shows that CEM improves as we increase the candidate set and is consistently better than greedy based algorithms. 

\vspace{-0.4cm}
\begin{center}
\begin{table}[!htb]
\hspace{0.3cm}\begin{tabular}{|p{1.5cm}|p{0.8cm}|p{1.5
cm}|p{1.4cm}|p{0.6cm}|}
\hline
\small Candidate Set Size $|\mathcal{G}|$ & \small Greedy & \small Greedy (refinement) & \small Beamsearch (width 5) & \small CEM \\
\hline\hline
\small 25 & \small 92.31 & \small 91.79 & \small 92.31 & \small 93.25 \\
\hline
\small 50 & \small 93.16 & \small 92.74 & \small 93.50 & \small 93.59 \\
\hline
\small 75 & \small 93.08 & \small 93.16 & \small 92.99 & \small 93.68 \\
\hline
\small 100 & \small 92.82 & \small 92.65 & \small 92.05 & \small 94.02 \\
\hline
\small 150 & \small 93.76 & \small 94.02 & \small 93.85 & \small 94.44\\
\hline
\end{tabular}
\caption{1-MDR - MCR ($\%$) for different algorithms with different candidate set size $|\mathcal{G}|$.}
\label{table:candidate_size}
\end{table}
\end{center}
\vspace{-0.4cm}

\vspace{-0.4cm}
\section{Conclusion and future work}
\label{sec:conclusion}

In this work, we focus on a small-footprint voice command recognition system composed of a CTC-based small-capacity acoustic encoder, and a corresponding maximum a posteriori decoder for the recognition of a  limited set of fixed commands. With a command specific dataset, we proposed a novel pipeline that automatically augments the command grammar for improved mis-detection and mis-classification rate. We achieved this by adapting the decoder to the consistent decoding variations of the acoustic model. An important direction of future work is to extend our grammar augmentation pipeline to provide personalization, i.e., to improve the recognition accuracy for a specific user by adapting the decoder to better fit both the AM and the user's pronunciation pattern. 
 
\newpage
\bibliographystyle{IEEEbib}
\bibliography{refs}

\end{document}